\documentclass[12pt]{article}
\usepackage{color}
\usepackage{cite}
\usepackage{times}
\usepackage{epsfig}
\usepackage{graphicx}
\usepackage{amsmath}
\usepackage{amssymb}

\usepackage[pagebackref=true,breaklinks=true,letterpaper=true,colorlinks,bookmarks=false]{hyperref}

\newcommand{\tr}{ {\rm tr }}
\newcommand{\Pcal}{ {\mathcal P }}

\newcommand{\OMIT}[1]{}

\title{Many-to-Many Graph Matching: \\
a Continuous Relaxation Approach}
\author{Mikhail Zaslavskiy\,$^{1,2,3,5}$, Francis Bach\,$^{4}$ and Jean-Philippe Vert\,$^{1,2,3}$\\
\footnotesize{$^{1}$Centre for Computational Biology, Mines ParisTech, Fontainebleau, France}\\
\footnotesize{$^{2}$Institut Curie and $^{3}$INSERM, U900, Paris, F-75248 France}\\
\footnotesize{$^{4}$INRIA-WILLOW project, Ecole Normale Sup{\'e}rieure, Paris, France}\\
\footnotesize{$^{5}$Centre for Mathematical Morphology, Mines ParisTech, Fontainebleau, France}\\
}
\date{10 March 2010}
\begin{document}
\maketitle

\begin{abstract}
Graphs provide an efficient tool for object representation in various computer vision applications. Once graph-based representations are constructed, an important question is how to compare graphs.  This problem is often formulated as a graph matching problem where one seeks a mapping between vertices of two graphs which optimally aligns their structure. In the classical formulation of graph matching, only one-to-one correspondences between vertices are considered. However, in many applications, graphs cannot be matched perfectly and it is more interesting to consider many-to-many correspondences where clusters of vertices in one graph are matched to clusters of vertices in the other graph. In this paper, we formulate the many-to-many graph matching problem as a discrete optimization problem and propose an approximate algorithm based on  a continuous relaxation of the combinatorial problem. We compare our method with other existing methods on several benchmark computer vision datasets. 
\end{abstract}

\section{Introduction}
\label{sec:oto_gm}

Graphs provide a convenient and efficient tool for object representation in various computer vision applications. An image or an object in an image can typically be represented by a segmentation, contours, shock graph, or interest points~(see, e.g.,~\cite{ponce}).
 Once a graph representation is chosen, a fundamental question that often arises is that of \emph{comparing} graphs in order to compare images or objects. In particular, it is important in many applications to assess quantitatively the similarity between graphs (e.g., for applications in supervised or unsupervised classification), and to detect similar parts between graphs (e.g., for identification of interesting patterns in the data).

Graph matching is one approach to perform these tasks. In graph matching, one tries to ``align'' two graphs by matching their vertices in such a way that most edges are conserved across matched vertices. Graph matching is useful both to assess the similarity between graphs (e.g., by checking how much the graphs differ after alignment), and to capture similar parts between graphs (e.g., by extracting connected sets of matched vertices). 
This graph matching framework has many applications in computer vision, e.g.,  to match 2D or 3D shapes~\cite{Belongie02Shape,Zheng2006Robust,Leordeanu2005Spectral}, or to match deformable objects~\cite{Duchenne2009Tensor}, where methods based on linear or projective transforms usually fail~\cite{Fischler1981Random}.

Classically, only one-to-one mappings are considered in graph matching. In other words, each vertex of the first graph can be matched to only one vertex of the second graph, and vice-versa\footnote{Note that with the introduction of dummy nodes, one may match a vertex of the first graph to \emph{up to one} vertex of the second graph (see, e.g.,~\cite{Belongie02Shape})}.
This problem can be formulated as a discrete optimization problem, where one wishes to find a one-to-one matching which maximizes the number of conserved edges after alignment. This problem is NP-hard for general graphs, and remains impossible to solve \emph{exactly} in practice for graphs with more than 30 vertices or so. Therefore much effort has been devoted to the development of approximate methods which are able to find a ``good'' solution in reasonable time. These methods can roughly be divided into two large classes.  The first group consists of various local optimization algorithms on the set of permutation matrices, including  $A^*$-Beam-search \cite{Neuhaus06Fast} and genetic algorithms.
The second group consists in solving a continuous relaxation of the discrete optimization problem, such as the $\ell_1$-relaxation \cite{Almohamad1993linear}, the Path algorithm  \cite{Zaslavskiy2008patha}, various spectral relaxations \cite{Umeyama1988eigendecomposition, Caelli2004eigenspace, Carcassoni2002Spectral,Cour2006Balanced,Leordeanu2005Spectral} or power methods~\cite{Duchenne2009Tensor}.

In practice, we are sometimes confronted with situations where the notion of one-to-one mapping is too restrictive, and where we would like to allow the possibility to match groups of vertices of the first graph to groups of vertices of the second graph. We call such a mapping \emph{many-to-many}. For instance, in computer vision, the same parts of the same object may be represented by different numbers of vertices depending on the noise in the image or on the choice of object view, and it could be relevant to match together groups of vertices that represent the same part. From an algorithmic point of view, this problem has been much less investigated than the one-to-one matching problem. Some one-to-one matching methods based on local optimization over the set of permutation matrices have been extended to many-to-many matching, e.g., by considering the possibility to merge vertices and edges in the course of optimization \cite{Berretti04Graph, Ambauen2003Graph}. Spectral methods have also been extended to deal with many-to-many matching by combining the idea of spectral decomposition of graph adjacency matrices with clustering methods \cite{Keselman03many-to-many,Caelli2004eigenspace}. However, while the spectral approach for one-to-one matching can be interpreted as a particular continuous relaxation of the discrete optimization problem \cite{Umeyama1988eigendecomposition}, this interpretation is lost in the extension to many-to-many matching. In fact, we are not aware of a proper formulation of the many-to-many graph matching problem as an optimization problem solved by relaxation techniques.

Our main contribution is to propose such a formulation of the many-to-many graph matching problem as a discrete optimization problem, which generalizes the usual formulation for one-to-one graph matching (Section~\ref{sec:mtm_gm}), and to present an approximate method based on a continuous relaxation of the problem (Section~\ref{sec:gradient}). 
The relaxed problem is not convex, and we solve it approximately with a conditional gradient method. We also study different ways to map back the continuous solution of the relaxed problem into a many-to-many matching. We present experimental evidence in Section~\ref{sec:experiments}, both on simulated and simple real data, that this formulation provides a significant advantage over other one-to-one or many-to-many matching approaches.

\section{Many-to-many graph matching as an optimization problem}
\label{sec:mtm_gm}

In this section we derive a formulation of the many-to-many graph matching problem as a discrete optimization problem. We start by recalling the classical expression of the one-to-one matching problem as an optimization problem. We then show how to extend the one-to-one formulation to the case of one-to-many matchings. Finally we describe how we can define many-to-many matchings via two many-to-one  mappings.

\vspace{0.1cm}
\noindent \textbf{One-to-one graph matching.}~~
Let $G$ and $H$ be two graphs with $N$ vertices (if the graphs have different numbers of vertices, we can always add dummy nodes with no connection to the smallest graph).
We also denote by $G$ and $H$ their respective adjacency matrices, i.e, square $\{0,1\}$-valued matrices of size $N \times N$ with element $(i,j)$ equal to $1$ if and only if there is an edge between vertex $i$ and vertex $j$.

 A one-to-one matching between $G$ and $H$ can formally be represented by a $N\times N$ permutation matrix $P$, where $P_{ij}=1$ if the $i$-th vertex of graph $G$ is matched to the $j$-th vertex of graph  $H$, and $P_{ij}=0$ otherwise. Denoting by $\| \cdot \|_F$ the Frobenius norm of matrices, defined as $\|A\|_F^2 = \tr A^\top  A =(\sum_i\sum_j A^2_{ij})$, we note that $\|G-PHP^\top \|_F^2$ is twice the number of edges which are not conserved in the matching defined by the permutation $P$. The one-to-one graph matching problem is therefore classically expressed as the following discrete optimization problem:
\begin{equation}
\!\!\!\!\!\!
\begin{split}
\min_{P} & \ ||G-PHP^\top ||_F^2 \mbox{ subject to }  P \in \Pcal_{oto},  \mbox{ with }\\ 
&\!\!\!\! \!\!\Pcal_{oto} \!=\! \{P\in\{0,1\}^{N\times N}\!,   P1_N \!= \! 1_N, P^\top 1_N\! =\! 1_N\},
\end{split}
\label{eq:oto_gm}
\end{equation}
where $1_N$ denotes the constant $N$-dimensional vector of all ones.
We note that $\Pcal_{oto}$ simply represents the set of permutation matrices. The convex hull of this set is the set of doubly stochastic matrices, where the the constraint
$P\in\{0,1\}^{N\times N}$ is replaced by $P\in[0,1]^{N\times N}$.

\vspace{0.1cm}
\noindent \textbf{From one-to-one to one-to-many.}~~
Suppose now that $G$ has less vertices than $H$, and that our goal is to find a matching that associates each vertex of $G$ with one or more vertices of $H$ in such a way that each vertex of $H$ is matched to a vertex of $G$. We call such a matching \emph{one-to-many} (or many-to-one). The problem of finding an optimal one-to-many matching can be formulated as minimizing the same criterion as (\ref{eq:oto_gm}) but modifying the optimization set as follows:
\begin{equation*} 
\begin{split}
\Pcal&_{otm}(N_G,N_H) =\{P\in\{0,1\}^{N_G\times N_H}, \\ 
&P1_{N_H}\leq k_{ {max}}1_{N_G},~P1_{N_H}\geq 1_{N_G}, ~P^\top 1_{N_G}=1_{N_H}\}\ ,
\end{split}
\end{equation*} 
where $N_G$ denotes the size of graph $G$, $N_H$ denotes the size of graph $H$, and $k_{\mbox{max}}$ denotes an optional upper bound on the number of vertices that can be matched to a single vertex. As opposed to the one-to-one matching case, each row sum of $P$ is allowed to be larger than one, and the non-zero elements of the $i$-th row of $P$ corresponds to the vertices of graph $H$ which are matched to the $i$-th vertex of $G$.

Most of existing continuous relaxation techniques may be adopted for one-to-many matching. For example, \cite{Cour2006Balanced} describes how spectral relaxation methods may be used in the case of one-to-many matching. Other techniques like convex relaxation \cite{Zaslavskiy2008patha} may be used as well since the convex hull of $\Pcal_{otm}$ is also obtained by relaxing the constraint $P\in\{0,1\}^{N\times N}$ to $P\in[0,1]^{N\times N}$.

\begin{figure}[lht]
\centering
\includegraphics[width=6cm]{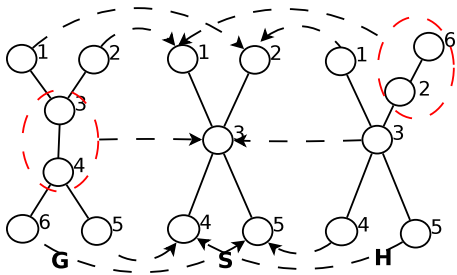}
\caption{Many-to-many matching between $G$ and $H$ via many-to-one matching of both graphs to a virtual graph $S$.}
\label{fig:mtm_gm}
\end{figure}

\vspace{0.1cm}
\noindent \textbf{From one-to-many to many-to-many.}~~
Now to match two graphs $G$ and $H$ under many-to-many constraints we proceed as if we matched these two graphs to a virtual graph~$S$ under many-to-one constraints, minimizing the difference between the transformed graph obtained from $G$ and the transformed  graph obtained from  $H$. The  idea of many-to-many matching as a double one-to-many matching is illustrated in Figure \ref{fig:mtm_gm}.
Graph $S$ (assumed to have $L$ vertices) represents the graph of matched vertex clusters. Each vertex of~$S$ corresponds to a group of vertices of $G$ and a group of vertices of $H$ matched to each other. Let $P_1\in \Pcal_{mto}(L,N_G) $ denote a many-to-one matching $G\rightarrow S$, and $P_2 \in \Pcal_{mto}(L,N_H)$ a many-to one matching $H\rightarrow S$; we propose to formulate the many-to-many graph matching problem as an optimization problem where we seek $S$, $P_1$ and $P_2$ which minimize the difference between $S$ and $P_1GP_1^\top $ and between $S$  and $P_2HP_2^\top $. The intermediate graph $S$ may be squeezed out by considering directly the difference between $P_1GP_1^\top $  and  $P_2HP_2^\top $. We end up with the following objective function for  the many-to-many graph matching problem:
\begin{equation}
F(P_1,P_2)=||P_1G_1^\top P_1^\top -P_2HP_2^\top ||^2_F,
\label{eq:f_mtm_gm}
\end{equation}
where $P_1 \in \Pcal_{mto}(L,N_G)$ and $P_2 \in \Pcal_{mto}(L,N_H)$ denote two many-to-one mappings. The objective function (\ref{eq:f_mtm_gm}) is similar to the objective function for the one-to-one case (\ref{eq:oto_gm}). In (\ref{eq:oto_gm}), we seek a permutation which makes the second graph $H$ as similar as possible to $G$.   In (\ref{eq:f_mtm_gm}), we seek \emph{combinations of merges and permutations} which makes $G$ and $H$ as similar as possible to each other. The only difference between both formulations is that in the many-to-many case we add the merging operation.

There are two slightly different ways of defining the set of matrices over which (\ref{eq:f_mtm_gm}) is minimized. We can fix in advance the number of matching clusters $L$, which corresponds to the size of $S$, in which case the optimization set is $P_1 \in \Pcal_{mto}(L,N_G)$ and $P_2 \in \Pcal_{mto}(L,N_H)$. An alternative way which we follow in the paper is to remove the constraint $P1_{N_G}\geq 1_{L}$ from the definition of $\Pcal_{mto}(L,N_G)$, in this case the method estimates itself the number of matching clusters (number of rows with non-zero sum).
Finally, we thus formulate the many-to-many graph matching problem as follows:
\begin{equation}
\begin{split}
&\min_{P_1,P_2}||P_1G_1^\top P_1^\top -P_2HP_2^\top ||^2_F \ \ \mbox{   subject to }\\
&P_1\in\{0,1\}^{N_K\times N_G},~P_11_{N_G}\leq k_{max}1_{N_K},~P_1^\top  1_{N_K}=1_{N_G},\\ 
&P_2\in\{0,1\}^{N_K\times N_H},~P_21_{N_H}\leq k_{max}1_{N_K},~P_2^\top 1_{N_K}=1_{N_H}\ , 
\end{split}
\label{eq:mtm_gm}
\end{equation}
where $N_K=\min(N_G,N_H)$ represents the maximal number of matching clusters.
This formulation is in fact valid for many kinds of graphs, in particular graphs may be directed (with asymmetric adjacency matrices), have edge weights (with real-valued adjacency matrices), and self-loops (with non-zero diagonal elements in the adjacency matrices). We also describe in Section \ref{sec:gradient} how this formulation can be modified to include information about vertex labels, which are important for computer vision (see, e.g., \cite{Belongie02Shape}). 

\section{Continuous relaxations of the many-to-many graph matching problem}
\label{sec:gradient}
The many-to-many graph matching problem (\ref{eq:mtm_gm}) is a hard discrete optimization problem. We therefore need an approximate method to solve it in practice. In this section we propose an algorithm based on a continuous relaxation of (\ref{eq:mtm_gm}). For that purpose we propose to replace the binary constraints $P_1\in\{0,1\}^{N_K\times N_G}$, $P_2\in\{0,1\}^{N_K\times N_H}$ by continuous constraints $P_1\in[0,1]^{N_K\times N_G}$, $P_2\in[0,1]^{N_K\times N_H}$. Note that if we had a linear objective function in $(P_1,P_2)$, the continuous relaxation would be exact because we simply replace the optimization set by its convex hull. However, our objective function (\ref{eq:mtm_gm}) is quartic, and its optimum is in general not an extreme point of the optimization set. 
To solve the relaxed optimization problem we propose to use the following version of the conditional gradient (a.k.a. Franck-Wolfe method \cite{Bertsekas1999Nonlinear}):
\begin{itemize}
\item Input: initial values $P_1^0$ and $P_2^0$, $t=0$,
\item Do
\begin{enumerate}
\item compute $\nabla F(P_1^t,P_2^t)$
\item find the minimum of $ \nabla F(P_1^t,P_2^t)^\top (P_1,P_2)$ w.r.t.~$(P_1,P_2)$
\item perform line search in the direction of the optimum found in Step 2, assign the result to $P_1^{t+1}$, $P_2^{t+1}$, $t=t+1$ 
\end{enumerate}
\item Until  $|\Delta F| + ||\Delta P_1||_F+||\Delta P_2||_F < \varepsilon$
\item Output: $P_1^t$, $P_2^t$.
\end{itemize}
The minimization of a linear function in step 2, i.e., $\min_P \nabla F(P_1^t,P_2^t)^\top (P_1,P_2)$ is a version of the linear semi-assignment problem, and reduces to the classical linear assignment problem by adding dummy nodes. We then have to solve a linear assignment problem for a $k_{max}(N_G+N_H)\times N_H$ matrix, which can be done efficiently by the Hungarian algorithm \cite{Kuhn1955Hungarian}. The solution of the line search step can be found in closed form since the objective function is a polynomial of the fourth  order.
 
The conditional descent algorithm converges to a stationary point of (\ref{eq:mtm_gm})~\cite{Bertsekas1999Nonlinear}.
Because of the non-convex nature of the objective function, we can only hope to reach a local minimum (or more generally a stationary point) and it is important to have a good initialization. In our experiments we found that a good choice is the fixed ``uniform'' initialization, where we initialize $P_1$ by $\frac{1}{N_K}1_{N_G}1_{N_H}^\top $ and $P_2$ by the identity matrix $I$. Another option would be to use a convex relaxation of one-to-one matching \cite{Zaslavskiy2008patha}.

Algorithm complexity is mainly defined by two parameters: $N=k_{max}(N_G+N_H)$ and $\varepsilon$. In general the number of iterations of the gradient descent scales as $O(\frac{\kappa}{\varepsilon})$ where $\kappa$ is the condition number of the Hessian matrix describing the objective function near a local minima \cite{Bertsekas1999Nonlinear}. $N$ has no direct influence on the number of iterations, but it defines the cost of one iteration, i.e., the complexity of the Hungarian algorithm $O(N^3)$.   

\vspace{0.1cm}
\noindent \textbf{Projection.}~~
Once we have reached a local optimum of the relaxed optimization problem, we still need to project $P_1$ and $P_2$ to the set of matrices with values in $\{0,1\}$ rather than in $[0,1]$. Several alternatives can be considered. A first idea is to use the columns of $P_1$ and $P_2$ to define a similarity measure between the vertices of both graphs, e.g., by computing the dot products between columns. Indeed, the more similar the columns corresponding to two vertices, the more likely these vertices are to be matched if they are from different graphs, or merged if they are from the same graph. Therefore a first strategy is to run a clustering algorithm (e.g., K-means or spectral clustering) on the column vectors of the concatenated matrix $(P_1,P_2)$ and then use the resulting clustering to construct the final many-to-many graph matching. 

 An alternative to clustering is  an incremental projection or forward selection projection, which uses the matching objective function at every step.
Once $P_1$ and $P_2$ are obtained from the continuous relaxation, we take the pair of vertices $(g,h)$ from the union of the graphs having the most similar column vectors in $(P_1,P_2)$. We then re-run the continuous relaxation with the new (linear) constraint that these two vertices remain matched. We then go on and find the most similar pair of vertices from the constrained continuous solution. This greedy scheme can be iterated until all vertices are matched.

In our experiments, the second approach produced better results. This is mainly due to the fact that when we just run a clustering algorithm we do not use the objective function, while when we use incremental projection we adapt column vectors of unmatched vertices according to earlier established matchings. 
 
\vspace{0.1cm}
\noindent \textbf{Neighbor merging.}~~
In many cases, it can be interesting to favor the merging of neighboring vertices, as opposed to merging of any sets of vertices. To that end we propose the following modification to (\ref{eq:mtm_gm}):
$$
F_N(P_1,P_2)=F(P_1,P_2)-\tr G^\top P_1^\top P_1-\tr H^\top P_2^\top P_2.
$$
The matrix product $P_1^\top P_1$ is a $N_G\times N_G$ matrix, with $(i,j)$-th entry equal to $1$ if $i$ and $j$ are merged into the same cluster. Therefore, the new components in the objective function  represent the number of pairs of adjacent vertices merged together in $G$ and $H$, respectively.

\vspace{0.1cm}
\noindent \textbf{Local similarities.}~~
Like the one-to-one formulation, we can easily modify the many-to-many graph matching formulation to include information on vertex pairwise similarities by modifying the objective function as follows:
\begin{equation}
\begin{split}
F_\lambda(P_1,P_2)=(1-\lambda) F(P_1,P_2) + \lambda \tr C^\top P_1^\top P_2\ ,
\end{split}
\label{eq:f_mtm_gm_lambda}
\end{equation}
where the matrix $C \in {\mathbb{R}}^{N_G\times N_H}$ is a matrix of local dissimilarities between graph vertices, and parameter $\lambda$ controls the relative impact of information on  graph vertices and information on graph structures. The new objective function is again a polynomial of the fourth order, so our algorithm may still be used directly without any additional modifications.

\section{Related methods} \label{sec:related}
There exist two major groups of methods for many-to-many graph matching, which we briefly describe in this section. The first one consists of local search algorithms, generally used in the context of the graph edit distance, while the second one is composed of variants of the spectral approach.

\vspace{0.1cm}
\noindent \textbf{Local search algorithms.}~~
\label{sec:beam}
Examples of this kind of approach are given in \cite{Berretti04Graph} and \cite{Ambauen2003Graph}. In the classical formulation of the graph edit distance, the set of graph edit operations consists of deletion, insertion and substitution of vertices and edges. Each operation has an associated cost, and the objective is to find a sequence of operations with the lowest total cost transforming one graph into another. In the case of many-to-many graph matching, this set of operations is completed by merging (and splitting if necessary) operations. Since the estimation of the optimal sequence is a hard combinatorial problem, approximate methods such as beam search \cite{Neuhaus06Fast} as well as other examples of best-first, breadth-first and depth-first searches are used. 

\vspace{0.1cm}
\noindent \textbf{Spectral approach.}~~
Caelli and Kosinov \cite{Caelli2004eigenspace} discuss how spectral matching may be used for  many-to-many graph matching. Their algorithm is similar to the Umeyama method \cite{Umeyama1988eigendecomposition}
 but instead of one-to-one correspondences, they search a many-to-many mapping by running a clustering algorithm. In the first step, the spectral decomposition of graph adjacency matrices is considered
\begin{equation}
G=V_G\Lambda_GV_G^\top ,~H=V_H\Lambda_HV_H^\top .
\end{equation}
Rows of eigenvector matrices $V_G$  and $V_H$ are interpreted as spectral coordinates of graph vertices. Then vertices having similar spectral coordinates are clustered together by a clustering algorithm, and vertices grouped in the same cluster are considered to be matched.

Another example of spectral approach is given in \cite{Keselman03many-to-many} where, roughly speaking, the adjacency matrix is replaced by the matrix of shortest path distances, and then spectral decomposition with further clustering is used.
 
\section{Experiments}
\label{sec:experiments}
In this section we compare the new method proposed in this paper with existing techniques (beam-search and spectral approach). We  thus test three competitive approaches on several experiments:  beam-search ``Beam'' (A*-beam search from \cite{Neuhaus06Fast}), the spectral approach ``Spec'' \cite{Caelli2004eigenspace}, and our new gradient descent method ``Grad'' (from Section~\ref{sec:gradient}). 

\subsection{Synthetic examples}
\label{sec:synt}
In this section, we compare the three many-to-many graph matching algorithms on pairs of randomly generated graphs with similar structures. We generate graphs according to the following procedure: (1) generate a random graph $G$ of size $N$, where each edge is present with probability $p$, (2) build  a randomly permuted copy  $H$ of $G$, (3) randomly split the vertices in $G$ (and in $H$) by taking a random vertex in $G$ (and in $H$) and split it into two vertices (operation repeated    $M$ times), (4) introduce noise by  adding/deleting $\sigma \times p \times N^2$ random edges in both graphs.

\begin{figure*}
\centering
\includegraphics[width=4.6cm]{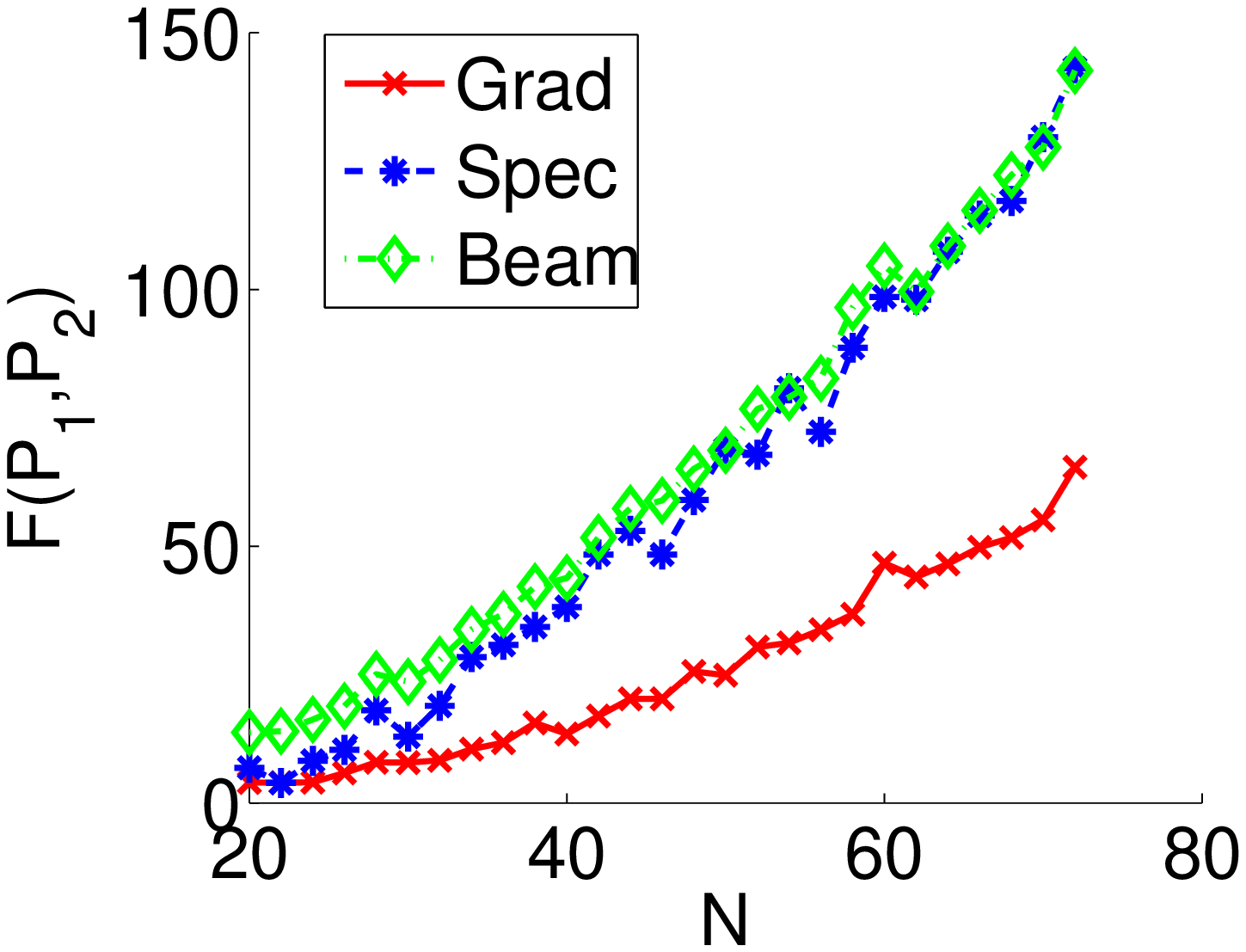}
\includegraphics[width=4.6cm]{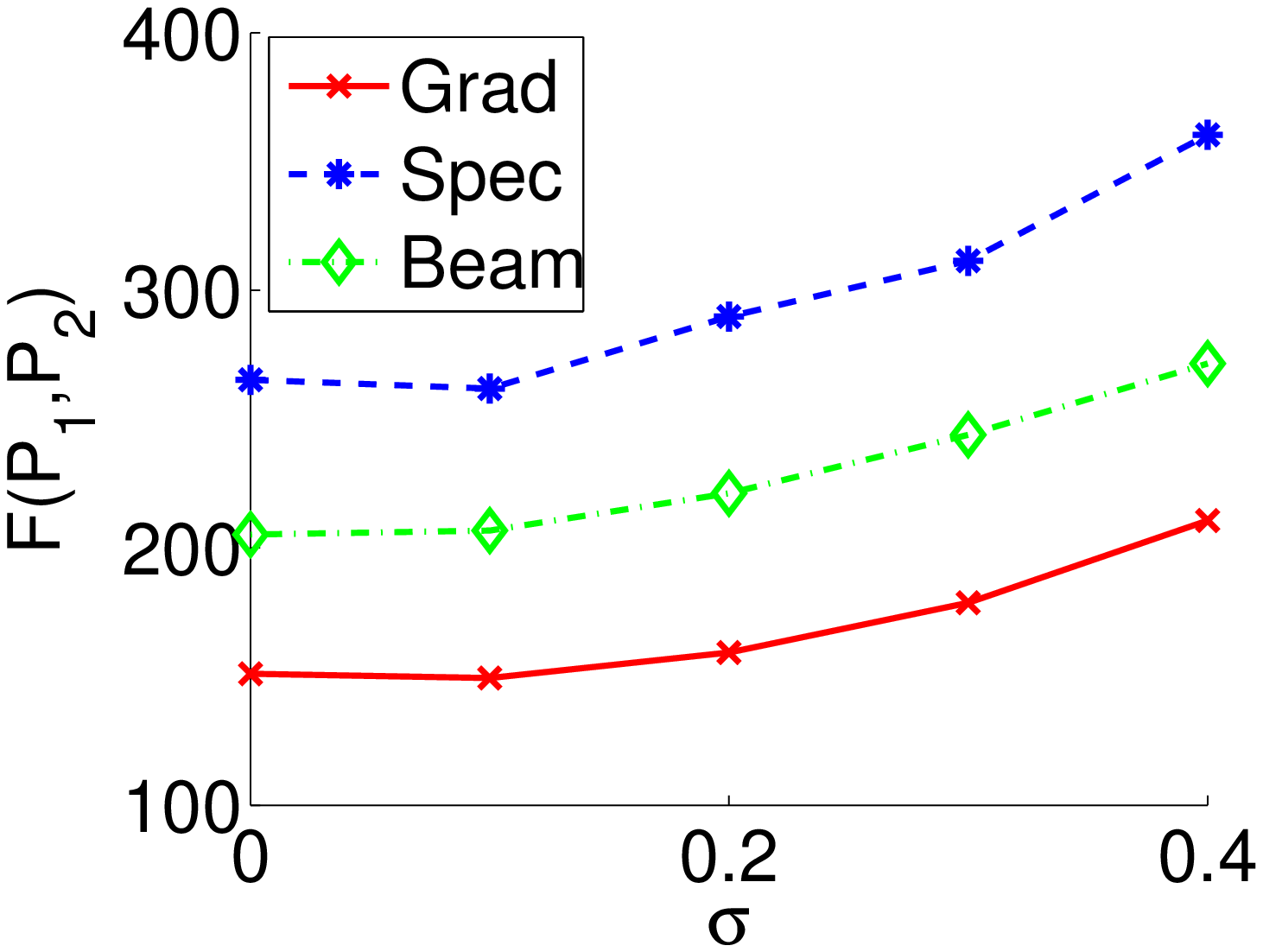}
\includegraphics[width=4.6cm]{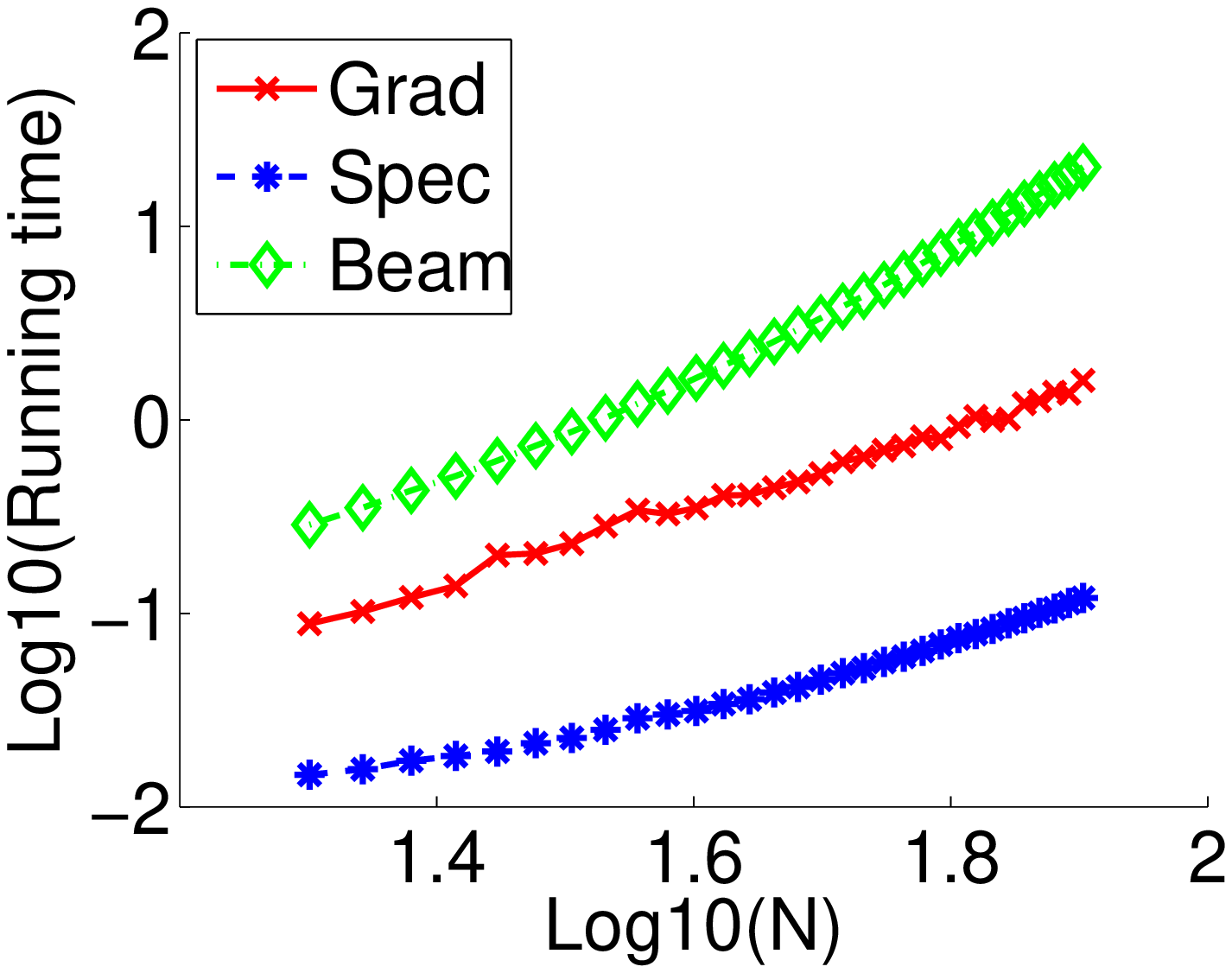}\\
\hspace{0.03\textwidth}(a) \hspace{0.23\textwidth} (b)\hspace{0.24\textwidth}(c)
\caption{(a) $F(P_1,P_2)$ (mean value over 30 repetitions) as a function of graph size $N$, simulation parameters: $p=0.1, \sigma=0.05, M=3$. (b) $F(P_1,P_2)$ (mean value over 30 repetitions) as a function of noise parameter $\sigma$, simulation parameters: $N=30, p=0.1, M=3$. (c) Algorithm running time (mean value over 30 repetitions) as a function of $N$ (log-log scale), other parameters are the same as in (a), ``Beam'' slope $\approx$ 3.8, ``Grad'' slope $\approx$ 2.5, ``Spec'' slope $\approx$ 2.7.}
\label{fig:num_experiments}
\end{figure*}

As already mentioned, our principal interest here is to understand the behavior of graph matching algorithms as a function of the graph size $N$, and their ability to resist to structural noise. Indeed, in practice we never have identical graphs and it is important to have a robust algorithm which is able to deal with noise in graph structures. The objective function $F(P_1,P_2)$ in (\ref{eq:mtm_gm}) represents the quality of graph matching, so to compare different graph matching algorithms we plot $F(P_1,P_2)$ as a function of $N$ (Figure \ref{fig:num_experiments}a), and $F(P_1,P_2)$ as a function of $\sigma$ (Figure \ref{fig:num_experiments}b) for the three algorithms. In both cases, we observe that ``Grad'' significantly outperforms both ``Beam''  and ``Spec''. ``Beam''  was run with beam width equal to 3, which represents a good trade-off between quality and complexity, ``Spec'' was run with projection on the first two eigenvectors with the normalization presented in \cite{Caelli2004eigenspace}\footnote{``Spec'' variants with three and more eigenvectors were also tested, but two eigenvectors produced almost the same matching quality and worked faster.}.   Figure \ref{fig:num_experiments}c shows how algorithms scale in time with the graph size $N$. The ``Spec'' algorithm is the fastest one, but ``Grad'' has the same complexity order  as ``Spec'' (corresponding curves  are almost parallel lines  in log-log scale, so both functions are polynomials with the same degree and different multiplication constants), these curves are coherent with theoretical values of algorithm complexity summarized in Section \ref{sec:gradient}. The ``Beam'' algorithm is much slower, and it also has worse complexity order.  
\subsection{Chinese characters}
\label{sec:chinese}
In this section we \emph{quantitatively} compare many-to-many graph matching algorithms as parts of a classification framework. We use graph matching algorithms to compute similarity/distance between objects of interest on the basis of their graph-based representations. As the classification problem, we chose the ETL9B dataset of Chinese characters. This dataset is well suited for our purposes, since Chinese characters may be naturally represented by graphs with variable non-trivial structures.

\begin{figure*}[htbp]
\centering
\raisebox{0cm}{\includegraphics[width=2.7cm]{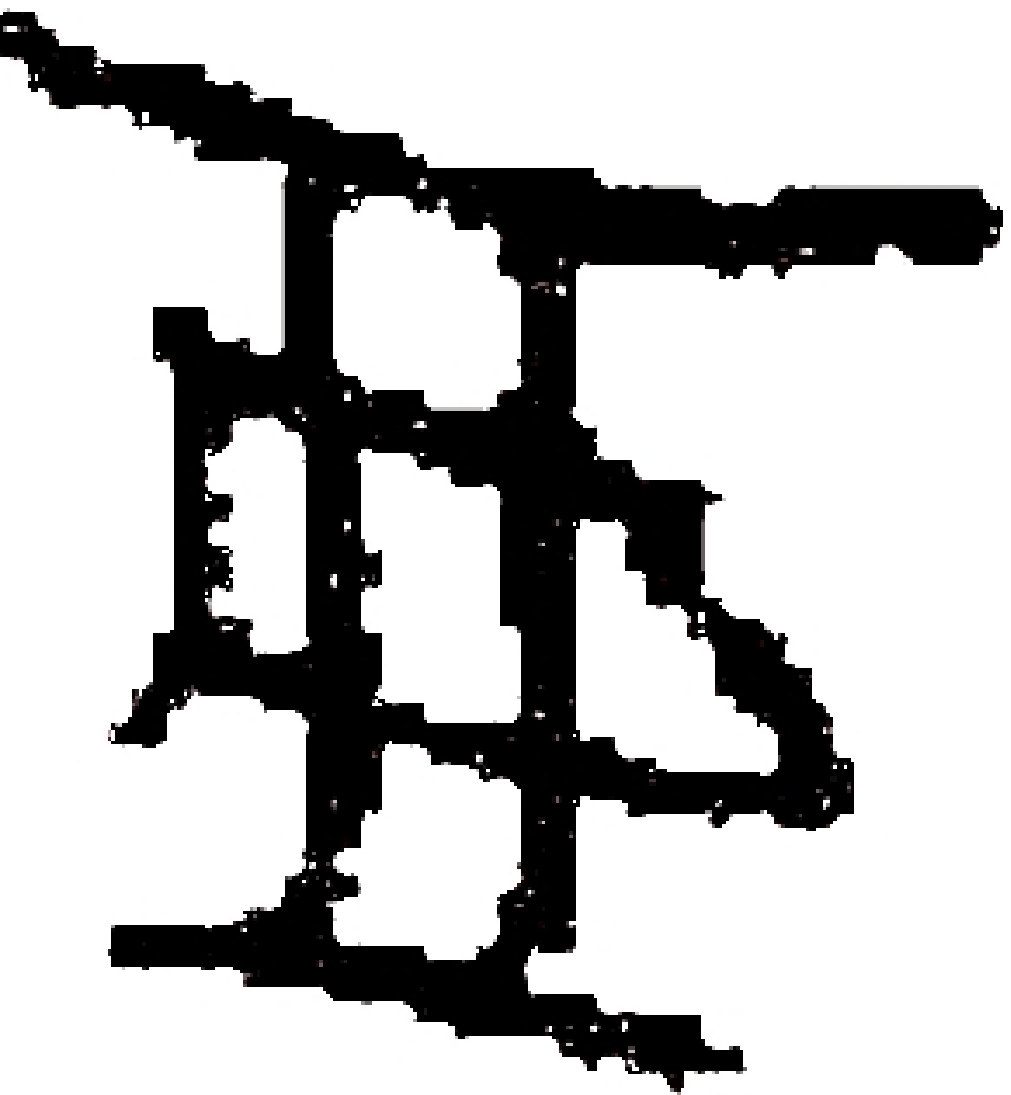}
\hspace{0.4cm}
\includegraphics[width=3cm]{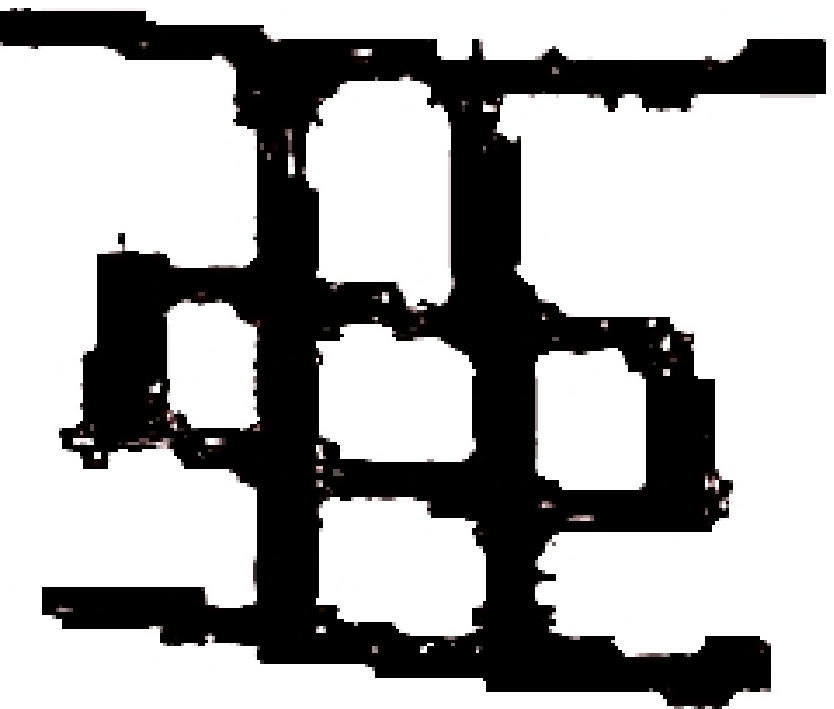}}
\raisebox{-0.8cm}{
\includegraphics[width=5.6cm]{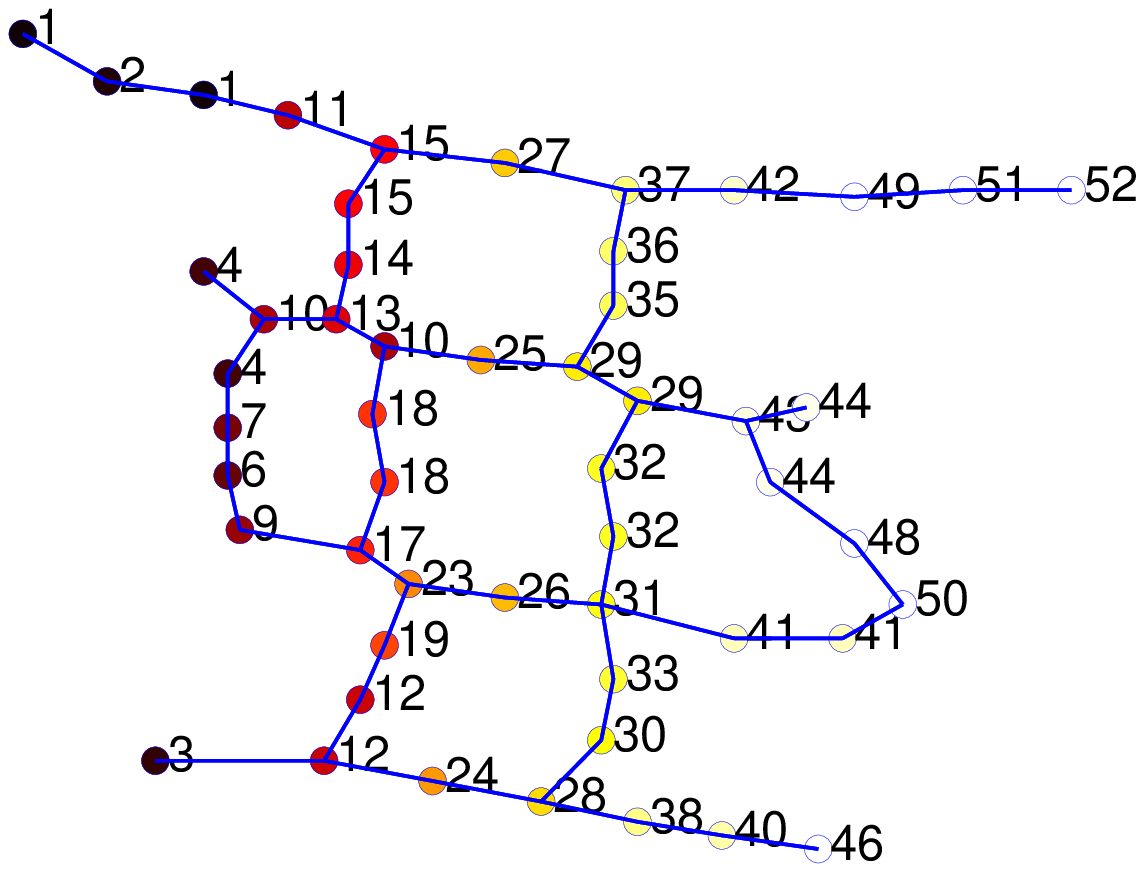}
\includegraphics[width=5cm]{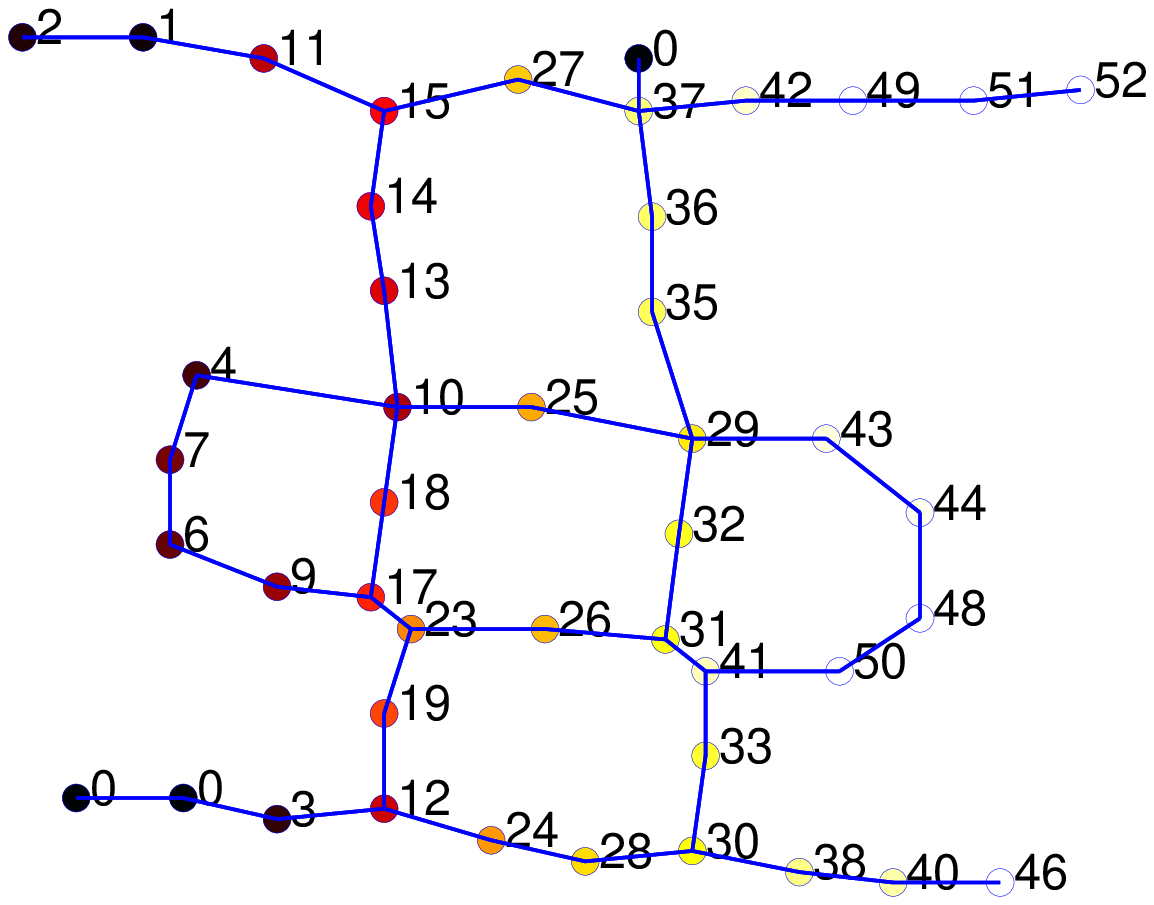}}
\caption{ Different writings of the same Chinese character and the matching of the corresponding graphs  made by ``Grad''. Vertices having the same id's are matched to each other.}
\label{fig:ch_chars}
\end{figure*}

Figure \ref{fig:ch_chars} illustrates how ``Grad'' works on graphs representing Chinese characters. We see that our algorithm produces a good matching, although not perfect, providing a correspondence between ``crucial'' vertices. 
The characters represented in Figure \ref{fig:ch_chars} are however very easy to recognize, and most classification algorithms show a good performance on them; for example, ``Grad'' produces a classification error rate below $0.2 \%$. 
To test graph matching algorithms on more challenging situations, we chose three ``hard to classify'' Chinese characters, i.e., three characters sharing similar graph structures, as illustrated in Table \ref{tab:ch_chars_hard}. We ran k-nearest neighbor (k-NN) with graph matching algorithms used as distance measures. The dataset consists of 600 images, 200 images of each class.

 Table \ref{tab:ch_chars_hard} shows classification results for the three many-to-many graph matching algorithms. In addition we report results for other popular approaches, namely, a SVM classifier with linear and Gaussian kernels, one-to-one matching with the Path algorithm (taken from \cite{Zaslavskiy2008patha}) and two versions of the shape context method \cite{Belongie02Shape}, with or without thin plate spline smoothing. The version named ``shape context'' computes polar histograms with further bipartite graph matching. To run the ``shape context+tps'' method we used code available online\footnote{{\tt http://www.eecs.berkeley.edu/vision/shape/}}.
  
Graph matching algorithms are run using information on vertex coordinates through (\ref{eq:f_mtm_gm}). The elements of the  matrix $C$ are defined as $C_{ij}=e^{-(x_i-x_j)^2-(y_i-y_j)^2}$. The parameter  $\lambda$ in (\ref{eq:f_mtm_gm}) as well as $k$ (number of neighbors in k-NN classifier) are learned via cross-validation.   
We see that the ``Grad'' algorithm shows the best performance, outperforming other many-to-many graph matching algorithms as well as other competitive approaches.
 
\begin{table}[htbp]
\caption{Top: chinese characters from three different classes. Bottom: classification results (mean and standard deviation of test error over cross-validation runs, with 50 repetitions of five folds)}
\label{tab:ch_chars_hard}
\centering
\includegraphics[width=2cm]{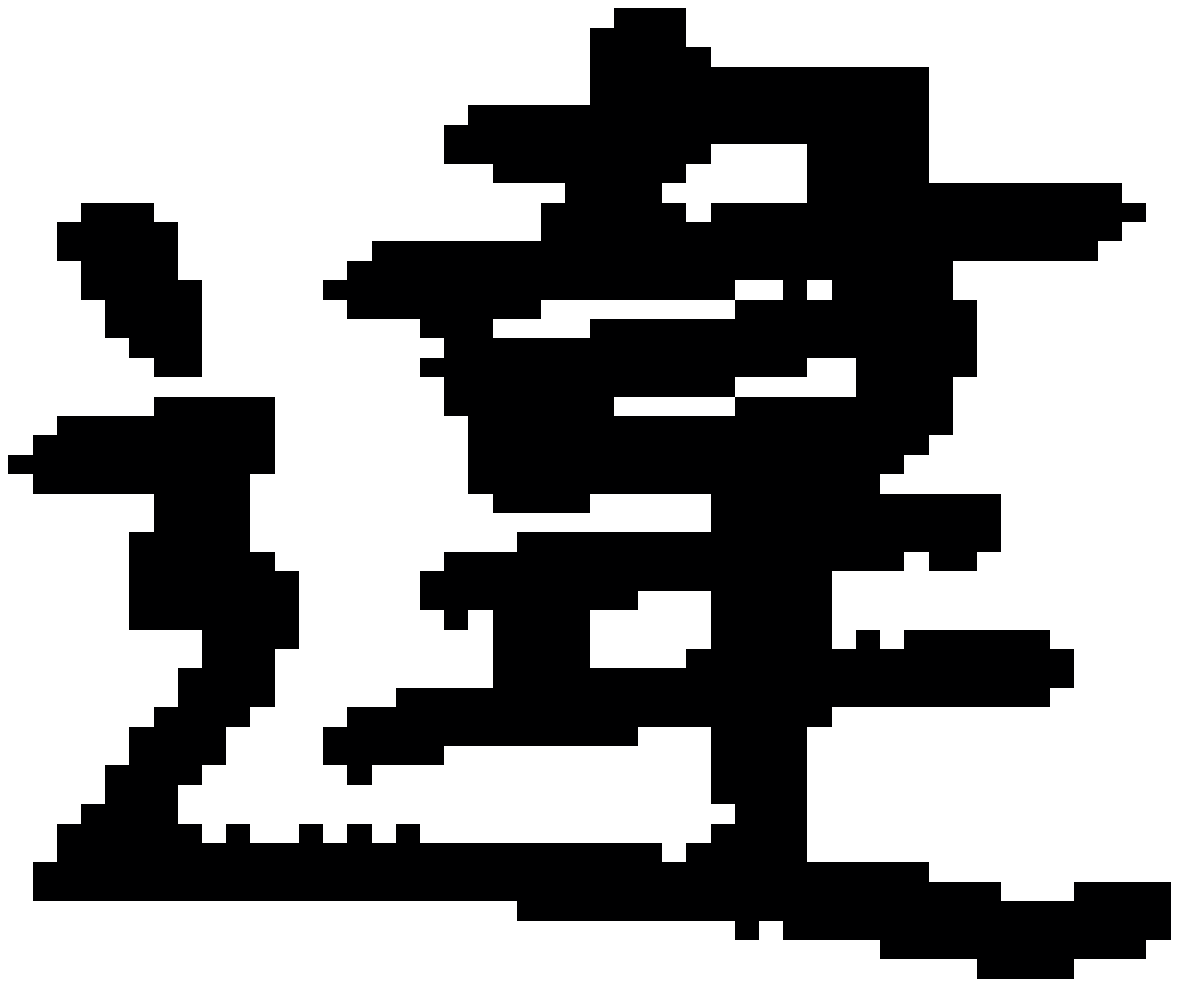}
\includegraphics[width=2cm]{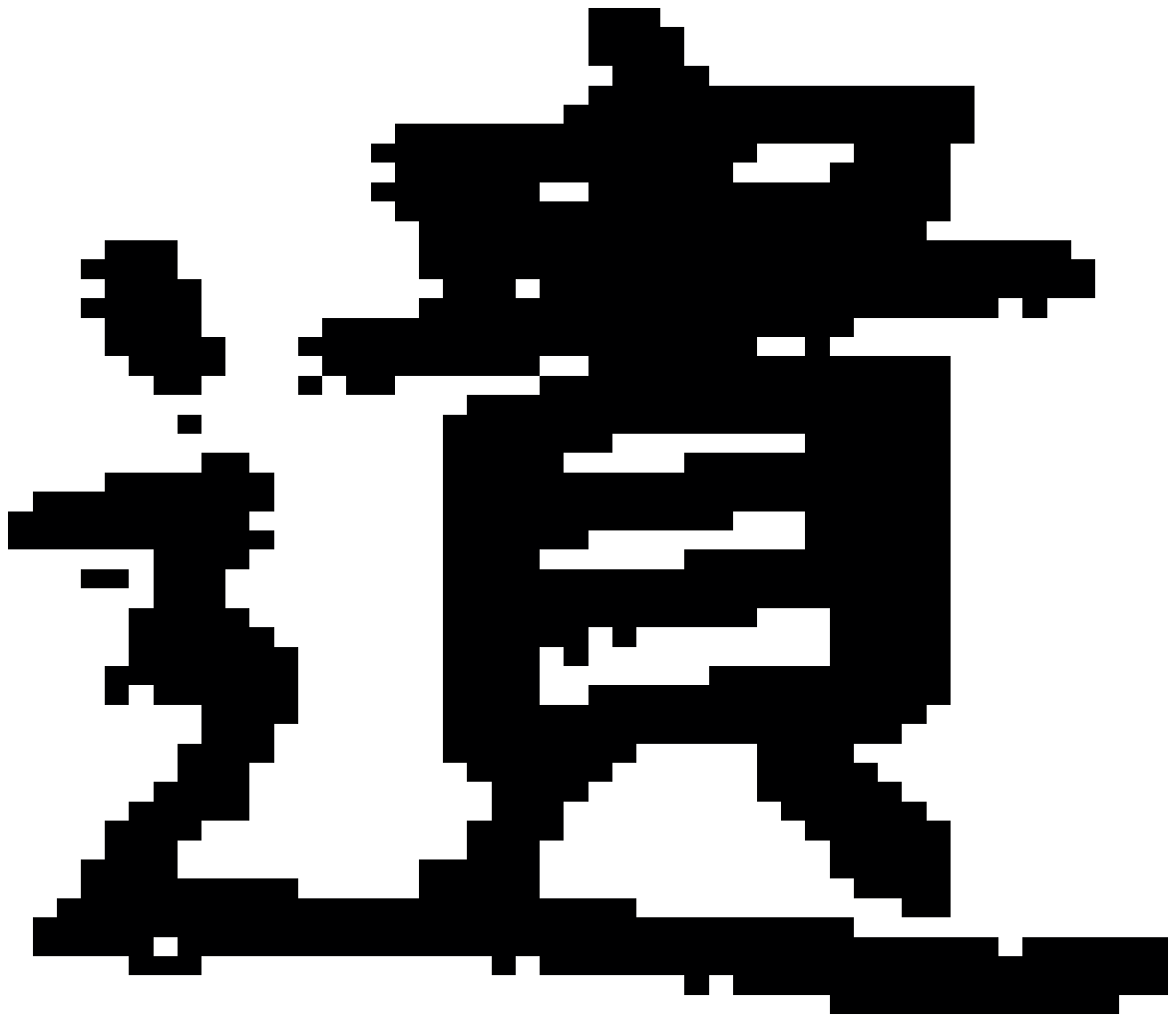}
\includegraphics[width=2cm]{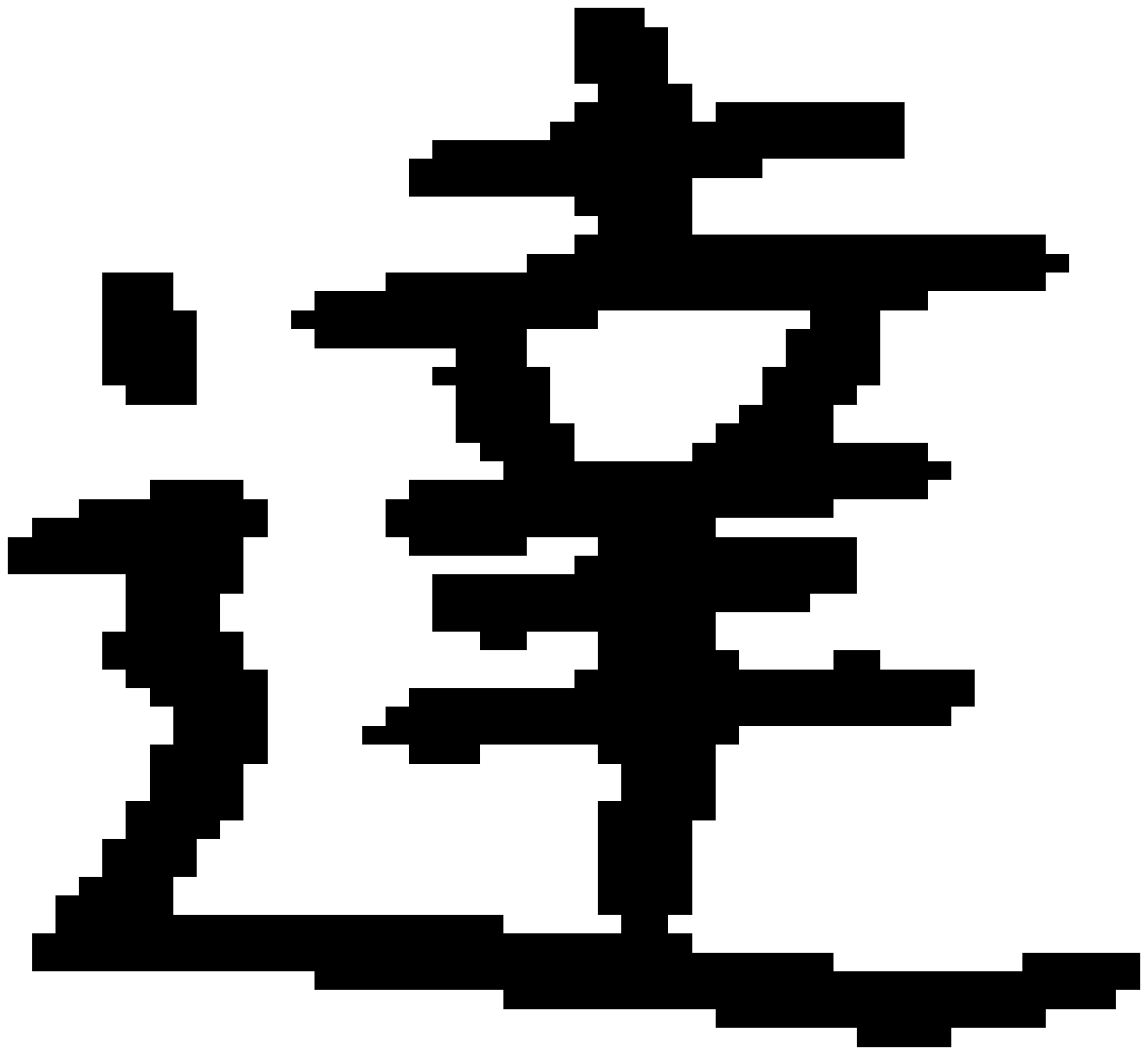}
\\
\begin{tabular}{|l|c|c|c|}
\hline
Method&${\rm error} $&${\rm STD} $\\
\hline
\hline
Linear SVM&0.377&$\pm$ 0.090 \\
SVM with Gaussian kernel&0.359&$\pm$ 0.076 \\
k-NN (one-to-one, Path) &0.248&$\pm$ 0.075 \\
k-NN (shape context) &0.399&$\pm$ 0.081 \\
k-NN (shape context+tps) &0.435&$\pm$ 0.092 \\
k-NN (Spec) &0.254&$\pm$ 0.071 \\
k-NN (Beam) &0.283&$\pm$ 0.079 \\
k-NN (Grad) &\textbf{0.191}&$\pm$ 0.063 \\
\hline
\end{tabular}
\end{table}

\subsection{Deformable objects matching}
\begin{figure*}[htbp]
\centering
\raisebox{1cm}{\includegraphics[width=6cm]{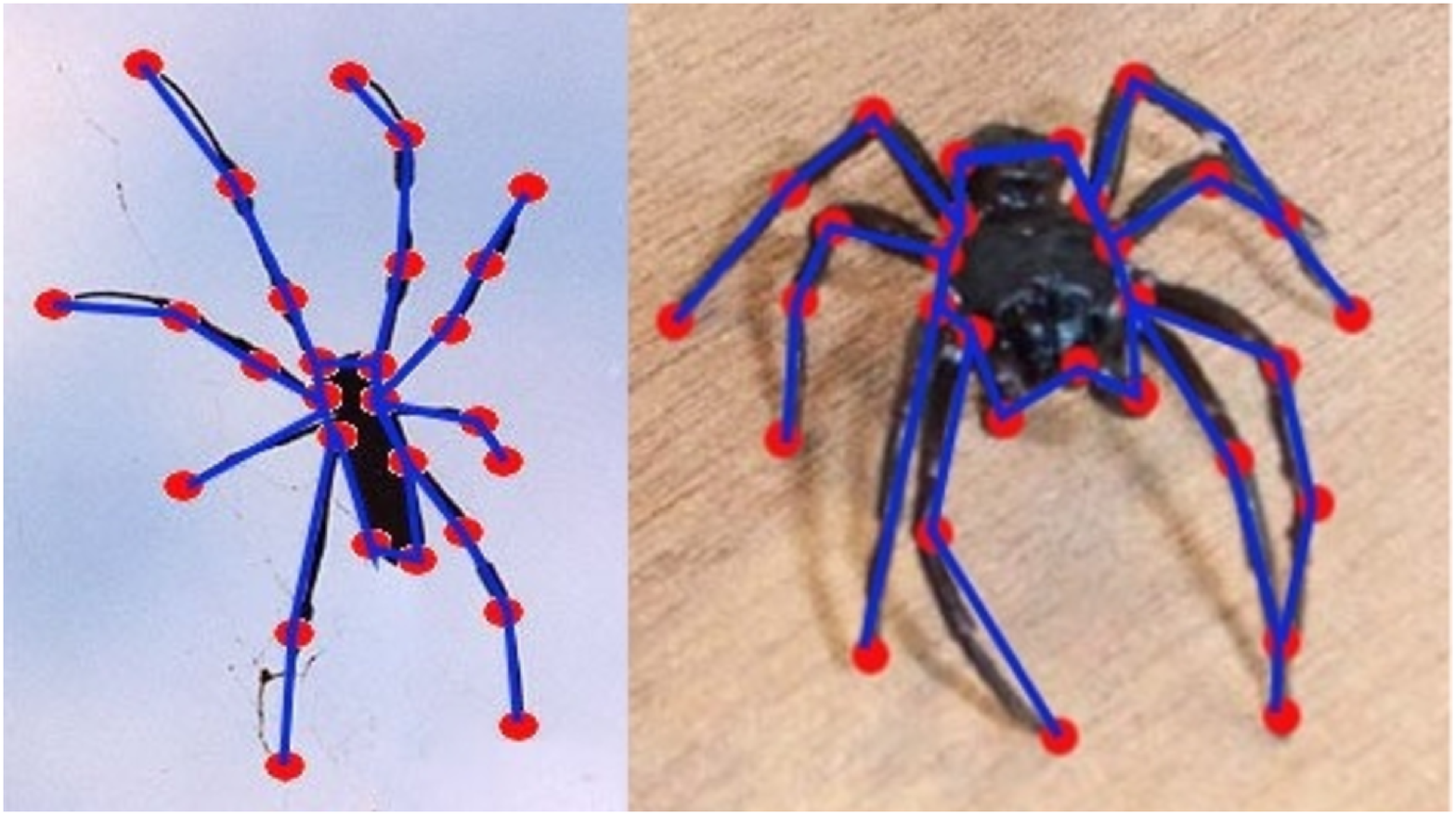}} \includegraphics[width=11cm]{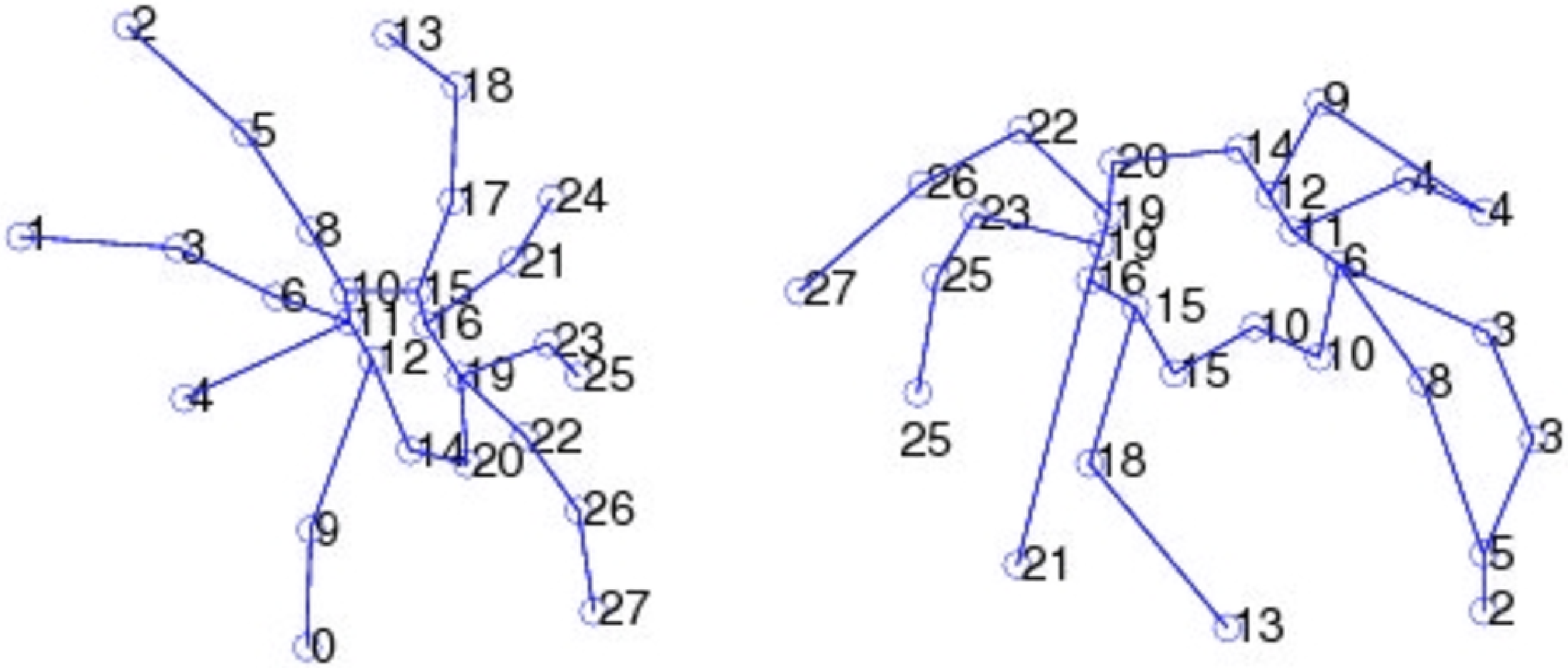}
\caption{Illustration of rotation invariant matching made by ``Grad''. Original spider photos with corresponding graph-based representations are given on the left. On the right, two spider graphs are aligned by ``Grad''.  Vertices with the same id's are matched to each other.}
\label{fig:spiders}
\end{figure*}

One advantage of graph-based image alignment algorithms is that they can be used in problems with deformable objects. Figure \ref{fig:spiders} shows how the ``Grad'' algorithm aligns a pair of photos of spiders (for which the graphs have been constructed by hand). These photos are taken from completely different viewpoints, which is a significant difficulty for many existing image alignment approaches based on the grouping of superpixels, such as \cite{Cour2007Recognizing, Tu2004Image}. Some methods generate various rotations or linear transforms of the same image and then take the best alignment~(see, e.g.,~\cite{Fischler1981Random,book}), but such approaches are not possible here because of deformations. Since image alignment should be rotation invariant we can not use the explicit vertex coordinates to construct the matrix $C$ as it was done in the previous section. Instead, we use the shape context features\cite{Belongie02Shape}; namely, each vertex gets a feature vector representing the polar histogram of the vectors joining this vertex to the other graph vertices. To make the polar histogram rotation invariant, we align the polar histograms by taking as an origin for angle the direction to the center of mass of all graph vertices. Under such a setup, polar histograms are invariant with respect to rotations around the graph center of mass.     

We see in Figure  \ref{fig:spiders} that ``Grad'' figures out that the top of the first image corresponds to the bottom of the second image, for example, it  groups two vertices representing the left part of the second spider head and matches them to one vertex of the left graph representing the same part in the first spider (vertices indexed by number 10).  

\subsection{Identification of object composite parts}
\label{sec:camels}
\begin{figure*}[htbp]
\centering
\includegraphics[width=3cm]{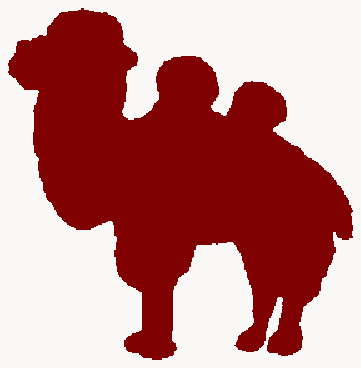}
\includegraphics[width=3cm]{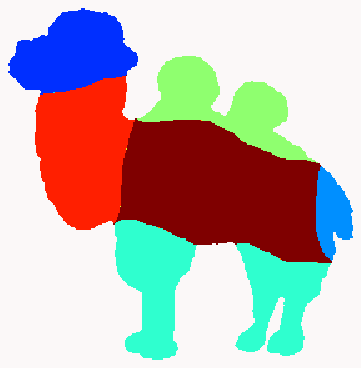}
\includegraphics[width=3cm]{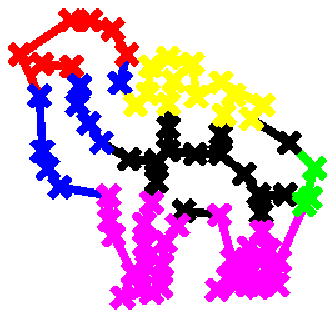}
\includegraphics[width=3cm]{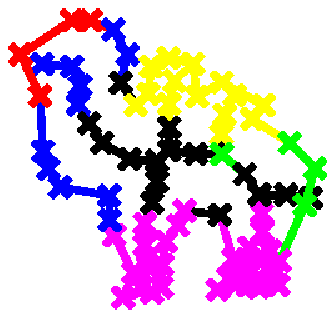}\\
\includegraphics[width=3cm]{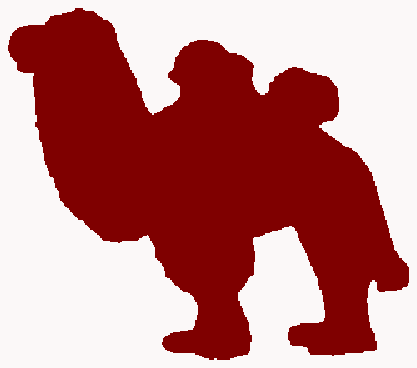}
\includegraphics[width=3cm]{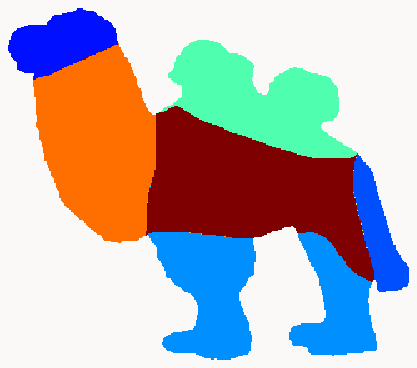}
\includegraphics[width=3cm]{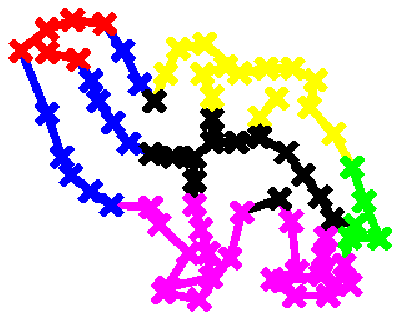}
\includegraphics[width=3cm]{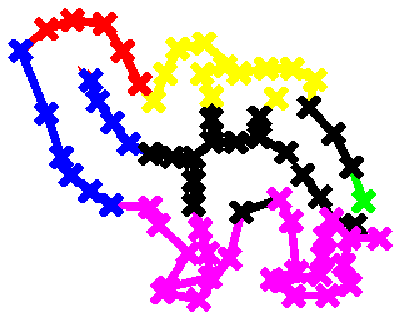}\\
(a)\qquad \qquad \qquad \qquad(b)\qquad \qquad \qquad \qquad(c)\qquad \qquad \qquad \qquad(d)
\caption{(a) Original images. (b) Manual segmentation (c) Graph-based representation (obtained automatically from subsampled contours and shock graphs) with induced vertex labels (d) Prediction of vertex labels on the basis of graph matching made by ``Grad''. Best seen in color.}
\label{fig:camels}
\end{figure*}

While the pattern recognition framework is interesting and important for the comparison of different graph matching algorithms, it evaluates only one aspect of these algorithms, namely, their ability to detect similar graphs. A second and important aspect  is their ability to correctly align vertices corresponding to the same parts of two objects. To test this capability, we performed the following series of experiments.  We chose ten camel images from the MPEG7 dataset and we divided by hand each image into 6 parts: head, neck, legs, back, tail and body (Figure \ref{fig:camels}). This image segmentation automatically defines a partitioning of the corresponding graph shown in the column (c) in Figure \ref{fig:camels}: all graph vertices are labeled according to the image part which they represent. Figure\ref{fig:camels} gives two illustrations of how   this procedure works. A good graph matching algorithm should map vertices from corresponding image parts to each other, i.e., heads to heads, legs to legs, and so on. Therefore to evaluate the matching quality of  the mapping, we use the following score. First, we match two graphs and then we try to predict vertex labels of one graph given the vertex labels of the second one. For instance, if vertex $g_1$ of the first image is matched to vertices $h_1$ and $h_2$ representing the head of the second image, then we predict that $g_1$ is of class ``head''.  The better the graph matching, the smaller the prediction error and vice-versa.

This experiment illustrates a promising application of graph matching algorithms. Usually segmentation algorithms extract image parts on the basis of different characteristics such as changing of color, narrowing of object form, etc. With our graph matching algorithm, we can extract segments which does not only have a specific appearance, but also have a semantic interpretation defined by a user (e.g., through the manual labelling of a particular instance).
 
Table \ref{tab:res_camels} presents mean prediction error over 45 pairs of camel images (we exclude comparison of identical images). Each pair has two associated scores: prediction error of the first image given the second one and vice-versa. We thus have 90 scores for each algorithm, which are used to compute means and standard deviations. Like in the previous sections, graph matching algorithms are run using information on vertex coordinates (using Eq.~(\ref{eq:f_mtm_gm})), with $C_{ij}=e^{-(x_i-x_j)^2-(y_i-y_j)^2}$. The parameter  $\lambda$ in (\ref{eq:f_mtm_gm}) as well as $k$ (number of neighbors in k-NN classifier) are learned via cross-validation. Here, again we observe that the ``Grad'' algorithm works better than other methods.
\begin{table}[htbp]
\caption{Identification of object composite parts: mean and standard deviation of prediction error (see text for details). Note that standard deviations are not divided by the square root of the sample size (therefore differences are statistically significant).}
\label{tab:res_camels}

\vspace*{.1cm}

\centering
\begin{tabular}{|l|c|c|c|c|}
\hline
&Grad&Spec&Beam&One-to-one\\
\hline
Error&\textbf{0.303}&0.351&0.432&0.342\\
\hline
STD&0.135&0.095&0.092&0.094\\
\hline
\end{tabular}
\end{table}
\section{Conclusion and Future work}
The main contribution of this paper is the new formulation of the many-to-many graph matching problem as a discrete optimization problem  and the approximate algorithm ``Grad'' based on a continuous relaxation.  The success of the proposed method compared to other competitive approaches may be explained by two reasons. First, methods based on continuous relaxations of discrete  optimization problems often show a better performance than local search algorithm due to their ability to better explore the optimization set with potentially large moves. Second, the ``Grad" algorithm aims to optimize a clear objective function naturally representing the quality of graph matching instead of a sequence of unrelated steps.

Besides a natural application of graph matching as a similarity measure between objects with complex structures, graph matching can also be used for object alignment. However, the structural noise usually encountered in graph-based representations have slightly hampered its application to natural images; but we believe that the many-to-many graph matching framework presented in this paper can provide an appropriate notion of robustness, which is necessary for computer vision applications. Of course, this requires the validation of our approach with graphs obtained from more cluttered images, which we are currently experimenting with.



\bibliographystyle{unsrt}
\bibliography{/home/michael/mysoft/jabref-2.4.2/bibli/bibli.bib}
\end{document}